\documentclass{article}
\usepackage{graphicx}
\newcommand{\myvspace}{\vspace{-0.5em}}
\PassOptionsToPackage{numbers, compress}{natbib} 
\usepackage{gensymb}

\usepackage{amsmath}

    \usepackage[preprint]{neurips_2024}

\usepackage[utf8]{inputenc} 
\usepackage[T1]{fontenc}    
\usepackage{hyperref}       
\usepackage{url}            
\usepackage{booktabs}       
\usepackage{amsfonts}       
\usepackage{nicefrac}       
\usepackage{microtype}      
\usepackage{xcolor}         

\title{Lung-DETR: Deformable Detection Transformer for Sparse Lung Nodule Anomaly Detection}

\author{
  Hooman Ramezani \\
  Department of Mechanical \& Industrial Engineering \\ 
  University of Toronto \\ 
  \texttt{hooman.ramezani@mail.utoronto.ca} 
  \And
  Dionne Aleman \\
  Department of Mechanical \& Industrial Engineering \\ 
  University of Toronto \\ 
  \texttt{dionne.aleman@utoronto.ca} 
  \And
  Daniel Létourneau \\
  Department of Radiation Oncology\\ 
  University of Toronto \\ 
  \texttt{daniel.letourneau@uhn.ca} 
}

\begin{document}
\maketitle
\begin{abstract}
Accurate lung nodule detection for computed tomography (CT) scan imagery is challenging in real-world settings due to the sparse occurrence of nodules and similarity to other anatomical structures. In a typical positive case, nodules may appear in as few as 3\% of CT slices, complicating detection. This paper presents a novel approach to lung tumor detection in CT data by framing the task as anomaly detection, targeting rare nodule occurrences in a predominantly normal dataset. Our novel method, named Lung-DETR combines Deformable Detection Transformer, Focal Loss, and Maximum Intensity Projection into a unified framework for sparse lung nodule detection. A 7.5mm Maximum Intensity Projection (MIP) is utilized to combine adjacent lung slices, decreasing nodule sparsity and enhancing spatial context to allow for better differentiation between nodules, bronchioles, and other complex vascular structures. Lung-DETR is trained with a custom focal loss function to better handle the imbalanced dataset, and outputs bounding boxes around detected nodules. Our model achieves an F1 score of 94.2\% (95.2\% recall, 93.3\% precision) on the LUNA16 dataset, with test dataset nodule sparsity of 4\% that is reflective of real-world clinical data.
\end{abstract}

\section{Introduction and Related Work}
Lung cancer remains one of the leading causes of cancer-related deaths globally; early detection is vital for improving patient outcomes. Despite significant advances in medical imaging, models see limited adoption in real-world settings. While there are many successful architectures for LUNA16 nodule detection that achieve high accuracy, many of the works include training on datasets of predominantly nodule-positive images. We fail to find a comprehensive solution that adequately addresses the issue of nodule sparsity in real-world data. For a model to be truly effective it must mitigate substantial class imbalance, where the number of slices containing only healthy tissue is much higher than those with lung nodules. The goal is to achieve high tumor detection accuracy while minimizing false positives and negatives. Such a model would be capable of providing meaningful medical insights to patients and could be deployed to underserved regions, offering affordable and accurate diagnoses for patients that could not otherwise access a physician.
\myvspace
\paragraph{Computed Tomography}(CT) data consists of volumetric images, created by concatenating cross-sectional slices of the body, which provide detailed views of internal structures. These slices are then stacked to form a comprehensive 3D representation of anatomical regions. However, nodule occurrences are sparse if they are present at all, with typically between 0 and to 3\% of slices showing signs of a nodule \cite{Walter2016Occurrence}. This imbalance presents a challenge for deep learning models, which must detect nodules while processing a disproportionately large volume of healthy slices.
\myvspace
\paragraph{Handling Imbalanced Data in Deep Learning} models is challenging because they are optimized to minimize overall error which leads to a bias favouring the majority class (e.g., healthy tissue) at the expense of the minority class (e.g., nodule slices). In cases of significant class imbalance, models are at risk of converging to a majority class classifier. This keeps error low and accuracy high but results in highly inaccurate detection of lung nodules. This issue is particularly critical in medical contexts, where false negatives, such as missed nodules, can have severe consequences. The scarcity of nodule data also hinders the model's ability to learn subtle distinctions necessary to differentiate nodules from other structures, further complicating detection.

There are various strategies to mitigate class imbalance, including oversampling, class weighting, and focal loss. Oversampling tumor slices artificially balances the dataset by increasing the number of minority class examples, but this approach misrepresents real-world conditions, leading the model to expect a higher prevalence of tumors than it will see at test time. Class weighting addresses imbalance by increasing the loss contribution of the minority class, forcing the model to pay more attention to underrepresented cases like tumors \cite{Buda2017A}. However, this can also increase false positives, as the model may overestimate the presence of the minority class \cite{Chan2019MetaFusion, Chan2020Controlled}. A more advanced approach, focal loss, modifies the cross-entropy loss by down-weighting well-classified examples (e.g., healthy slices) and emphasizing hard-to-classify ones like tumors, adjusting the loss based on prediction confidence. This method effectively targets the imbalance by prioritizing difficult examples, avoiding the shortcomings of class weighting and reducing the false positive rate, leading to improved precision and recall for rare classes \cite{Lin2017Focal}.
\myvspace
\paragraph{The LUNA16 Dataset} consists of 888 CT scan sets containing 1186 lung nodules. Lung nodules in LUNA16 are annotated based on the consensus of at least three out of four radiologists, with only nodules larger than 3 mm included as relevant findings. Nodules under 3 mm or identified by fewer than three radiologists are excluded from evaluation. LUNA16, derived from the LIDC-IDRI dataset, serves as a critical benchmark for developing deep learning models for lung nodule detection. Numerous studies using architectures such as CNNs, 3D-CNNs, U-Net, SAM, and V-Net have shown high detection accuracies on this dataset \cite{El-bana2020A, Gu2018Automatic}. However, variations in data processing across studies complicate direct comparison. These studies often focus on detecting nodules in slices already known to contain tumors, a task not reflective of real-world applications \cite{Xiao2020Segmentation}. Moreover, individual slices often lack the 3D context needed to differentiate between nodules and other structures, making it essential to incorporate adjacent slices in the analysis.
\myvspace
\paragraph{Maximum Intensity Projection} (MIP) enhances nodule visibility by combining adjacent CT into a single 2D image, projecting the highest-intensity voxels from adjacent slices to preserve crucial 3D spatial information. Widely used by radiologists, MIP helps distinguish nodules from vessels, vascular structures and bronchioles. Nodules generally appear as compact blobs, whereas vessels are elongated tube-like structures. This method is shown to be extremely effective in detecting  small pulmonary nodules between 3 mm and 10 mm while also reducing false positives \cite{Gruden2002Incremental, Zheng2019Automatic}.
\myvspace
\paragraph{Detection Transformer (DETR)}
Transformer architectures have become a strong alternative to CNNs in medical computer vision. While CNNs capture local features, they struggle with long-range dependencies, which refer to the model's ability to understand relationships between distant parts of an image, such as recognizing that a pattern in one corner of the scan may relate to another feature far across the image. This limitation arises because CNNs have a restricted receptive field, meaning they primarily focus on nearby pixels without fully capturing global context. Transformers use self-attention to capture complex relationships across the entire image. This is crucial for differentiating between structures such as nodules and vessels. DETR performs object detection by using self-attention to directly predict object locations \cite{Carion2020End-to-End}. However, DETR struggles with slow convergence and detecting small objects, such nodules \cite{Zhu2020Deformable}. Deformable-DETR improves efficiency by incorporating a custom attention mechanism that selectively focuses on a sparse set of relevant sampling points around a reference point, rather than attending to the entire feature map. This approach allows the model to dynamically adapt its focus to the most informative regions, enhancing efficiency and performance for small features such as  nodules in CT scans \cite{Zhu2020Deformable}.
\myvspace
\paragraph{Method Overview} This paper presents a novel approach to lung tumor detection in CT data by framing the task as anomaly detection with a focus on real-world applicability. Our method is the first to combine Deformable-DETR, Focal Loss, and Maximum Intensity Projection (MIP) into a unified framework specifically tailored for detecting sparse lung nodules. We build a customized transformer the training regimen for the processed LUNA16 dataset to address severe class imbalance by focusing the model's learning on difficult cases. This combination of architectural choices and training strategies has not been explored before in this context, allowing our model to achieve high sensitivity and precision in clinically relevant scenarios.

\section{Methodology}
\label{gen_inst}
\begin{figure}
    \centering
\includegraphics[width=1\linewidth]{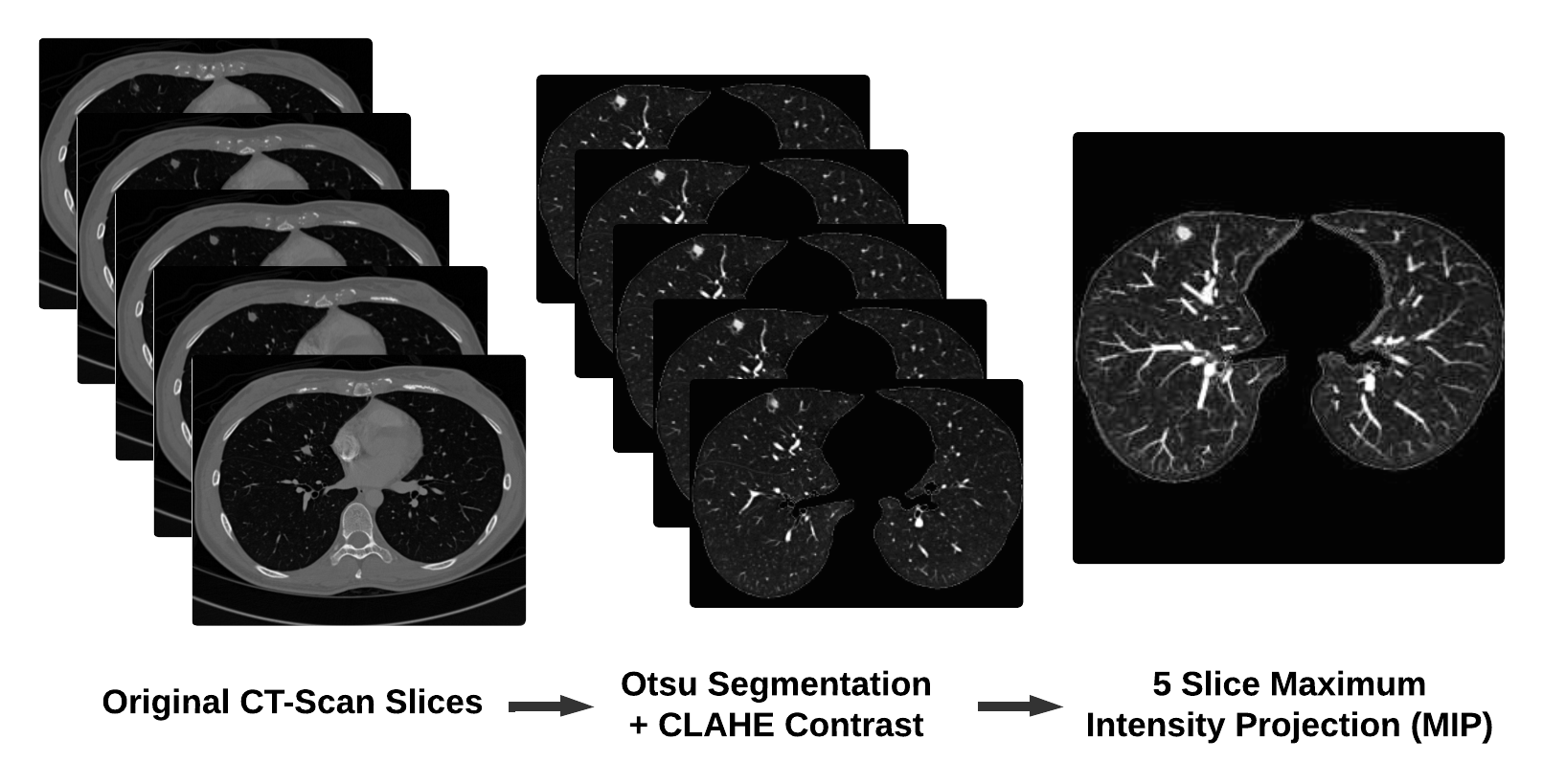}
    \caption{Data Processing Pipeline With Tumor Visible Top Left of Lung}
    \label{fig:enter-label}
\end{figure}
In this section, we describe our proposed approach for detecting sparse lung nodules in CT scans using Deformable-DETR, evaluated on the LUNA16 dataset. We train Deformable-DETR to achieve a balance between high sensitivity and specificity, detecting nodules in a dataset where healthy tissue dominates while minimizing false positives and negatives. Our custom LUNA16 preprocessing pipeline begins with isolating lung regions using Otsu's method for segmentation, followed by applying CLAHE to enhance contrast and direct the model's attention to the most relevant areas. Maximum Intensity Projection (MIP) is employed to merge adjacent CT slices into a single 2D image. To further optimize detection, we integrate a custom loss function that combines focal loss with the DETR loss function. The details of each component are described in the following subsections.
\subsection{Data Preprocessing}
Our preprocessing pipeline prepares CT scan data from the LUNA16 dataset for input into DETR, enhancing critical features and reducing noise. We visualize this process in Figure 1. CT data and mask annotations are loaded in MetaImage (mhd/raw) format. To standardize anatomical structures, images are resampled by calculating a resize factor based on the original and target voxel spacings, addressing inconsistencies between scans. The resampling factor \( R \) is calculated as shown in Equation \eqref{eq:resample}, where the image is scaled accordingly to achieve the desired voxel spacing:

\begin{equation}
R = \frac{S}{S'} = \left[\frac{S_x}{S'_x}, \frac{S_y}{S'_y}, \frac{S_z}{S'_z}\right]
\label{eq:resample}
\end{equation}

Otsu’s method is an image thresholding technique that automatically determines the optimal threshold value to separate foreground from background by minimizing intra-class variance. To reduce information, we utilize Otsu’s method to set a threshold that segments lung tissue from surrounding background structures to isolate the lung areas. This is followed by morphological operations, including connected component analysis and region erosion, to obtain clean binary masks to separate lungs from other features. Slices near the periphery, which provide minimal diagnostic information, are also automatically removed based on the size of the non-zero area. These steps decrease the number of non-zero pixels from around 15 million to 5.25 million per patient on average, allowing the model to focus on the most critical anatomical structures. After segmentation, we enhance contrast using Contrast Limited Adaptive Histogram Equalization (CLAHE), improving the visibility of subtle features like small nodules by adjusting contrast in localized regions. This is particularly useful when evaluating medical images as low contrast can obscure early-stage nodules and increasing can help models detect subtle abnormalities more effectively \cite{Sundaram2011Histogram}. This process is illustrated by the leftmost arrow of Figure 1.

Maximum Intensity Projection (MIP) projects the highest attenuation voxel from a 3D volume onto a 2D image \cite{Cody2002AAPM/RSNA}. This process can be mathematically described by Equation \eqref{eq:mip}, where the highest intensity voxel along the z-axis is selected for each (x, y) coordinate, producing a 2D image that highlights the most dense features of the volume. Based on empirical testing, a slab thickness of 7.5mm was found to best highlight nodules without surrounding structures. This process is illustrated by the rightmost arrow of Figure 1.
\begin{equation}
I_{\text{MIP}}(x, y) = \max_{z} \{ I(x, y, z) \}
\label{eq:mip}
\end{equation}

\subsection{Dataset}
The final processed dataset consists of 9,676 CT scan slices, each with a 7.5mm Maximum Intensity Projection (MIP) applied. Among these, 1,226 images are annotated with nodules, while the remaining 8,450 images contain healthy tissue. The dataset was split into 70\% for training, 20\% for validation, and 10\% for testing prior to any augmentation to avoid data contamination and ensure rigorous evaluation. In the training and validation sets, 12.7\% of the images contained a lung nodule. To better mimic real-world conditions, the test set had a reduced lung nodule rate of 3\%, contrasting with the higher rate used during training. This elevated rate in training was necessary to strike a balance between realism and model performance, as lower rates resulted in a dataset too sparse for effective training. Empirical tests confirmed that models trained on this higher rate generalized well when exposed to the lower nodule sparsity in testing. Post-split, a set of data augmentations was applied to the training set only to increase the dataset's size and variability. These include horizontal and vertical flips, rotations between -15\degree{} and +15\degree{}, brightness adjustments within -15\% to +15\%, and Gaussian noise (0.001 to 0.18\% SD) simulated typical CT scan sensor noise.

\subsection{Deformable-Detection Transformer}
\begin{figure}
    \centering
    \includegraphics[width=1\linewidth]{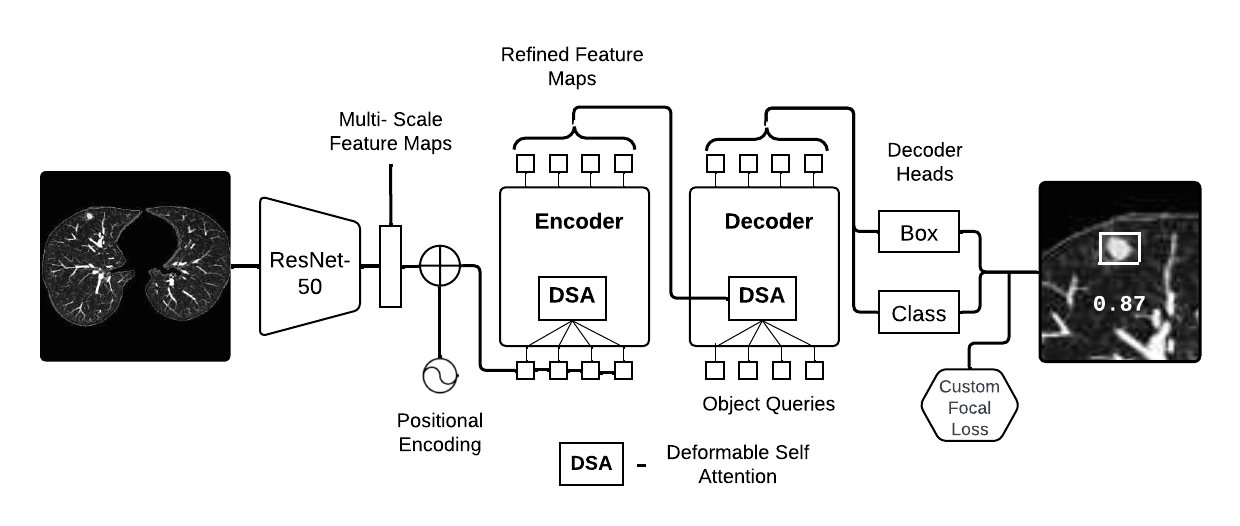}
    \caption{Lung-DETR Architecture}
    \label{fig:Lung-DETR}
\end{figure}
Detection Transformer (DETR) was chosen for lung tumor detection due to its strong performance in complex object detection tasks. To further enhance these capabilities, we adopted the deformable variant of DETR, as introduced by Zhu et al. \cite{Zhu2020Deformable}. Deformable attention dynamically focuses on a sparse set of sampling points around a reference point, making it both spatially adaptive and computationally efficient. By directing attention to the most relevant regions, Deformable-DETR significantly improves detection accuracy while reducing unnecessary computations and accelerating convergence.

Initial experimentation with DETR yielded a recall rate of 42\% after 20 epochs, performing well on tumors larger than 10mm but struggling with smaller ones. Switching to Deformable-DETR improved recall to over 80\% across all tumor sizes after just 8 epochs. With 74\% of tumors in the LUNA16 dataset measuring 3-10mm, the deformable attention variant was selected for tumor detection.

Figure 2 illustrates the custom deformable-DETR architecture used for sparse lung nodule detection. The detection task is formulated as a bounding box region proposal problem, where the model predicts bounding boxes and class probabilities for potential tumor regions. These predictions are evaluated against the ground truth annotations using an Intersection over Union (IoU) threshold of 50\%.

The proposed architecture begins by feeding processed Maximum Intensity Projection (MIP) images into a pretrained ResNet-50 backbone. This CNN backbone extracts multi-scale feature maps from stages C3 to C5 of ResNet-50, capturing both low-level textures and high-level semantic features to highlight critical lung regions. These feature maps are augmented by 2D sine-cosine positional encodings, which are crucial for preserving spatial relationships in 2D medical images, thus providing necessary spatial context to the encoder for accurate tumor detection.

The encoder utilizes a series of Deformable Self-Attention (DSA) layers to dynamically refine the multi-scale feature maps. Each DSA layer selectively attends to a sparse set of learnable sampling points around each query. The computational complexity of self-attention is \(O(H^2W^2C)\), where \(H\) and \(W\) are the feature map height and width, and \(C\) represents the number of channels. The encoder also integrates a multi-scale attention mechanism to process information at different feature scales, enhancing the model’s ability to detect nodules of varying sizes. The encoder outputs refined multi-scale feature maps enriched with context-aware representations.

The decoder stage consists of both cross-attention and self-attention modules. It starts by integrating the encoder’s refined feature maps with object queries, a learnable set of positional embeddings representing potential nodules within the image. The cross-attention modules leverage these object queries to dynamically interact with the encoder’s feature maps. This approach ensures the decoder directs attention efficiently to search for nodules, optimizing the detection of small nodules amidst complex lung structures.

The pipeline concludes with the decoder outputs being processed by two heads: the Bounding Box Regression Head, which predicts the coordinates (center, width, height) of potential nodules, and the Classification Head, which estimates the probability of each bounding box containing a nodule versus background. Both heads utilize the decoder’s output embeddings, with the regression head ensuring precise localization and the classification head accurately distinguishing nodules.

\subsection{Focal Loss for Classification}

To handle the significant class imbalance in the LUNA16 dataset, we customize the DETR loss function to incorporate focal loss. By adding a modulating factor, focal loss down-weights well-classified samples and emphasizes hard-to-classify samples, assisting in the detection of rare nodule instances. The focal loss function is defined in Equation \eqref{eq:focal_loss}:

\begin{equation}
FL(p_t) = - \alpha_t (1 - p_t)^\gamma \log(p_t),
\label{eq:focal_loss}
\end{equation}

where \( p_t \) is the predicted probability of the correct class, \( \alpha_t \) balances positive and negative examples, and \( \gamma \) adjusts focus towards challenging samples.

Empirical analysis demonstrated \( \gamma = 2 \) and \( \alpha_t = 0.25 \)  effectively balances the model's focus on hard-to-classify examples, improving detection of small pulmonary nodules. These values optimize the trade-off between precision and recall, minimizing false positives and negatives.

\begin{figure}
    \centering
    \includegraphics[width=1\linewidth]{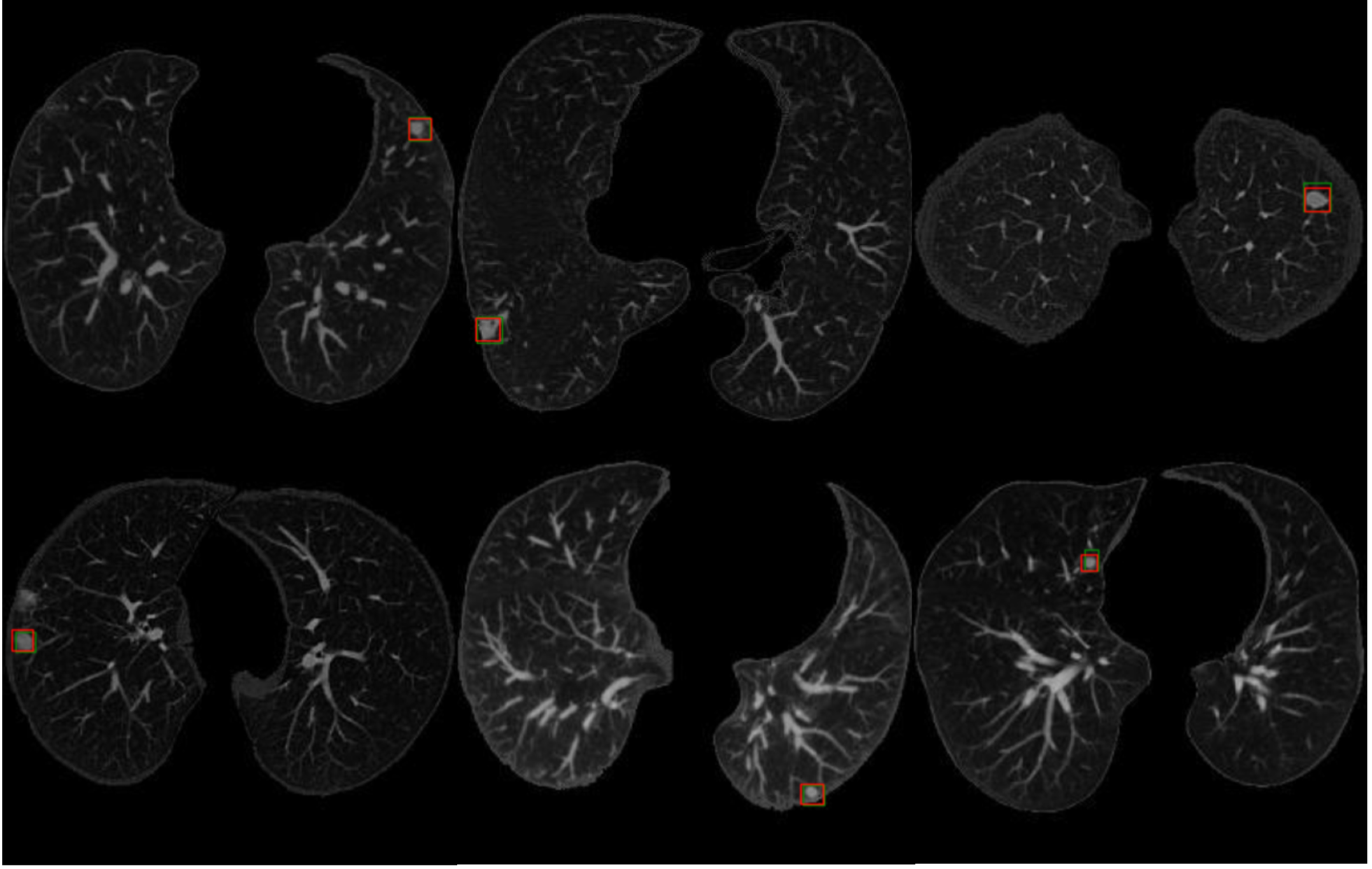}
    \caption{Lung-DETR Predicitions on Slices with Tumor}
    \label{fig:predvis}
\end{figure}
\section{Results}
This section evaluates the performance of the proposed Lung-DETR architecture on the LUNA16 dataset with a focus on key metrics such as recall, precision, and F1 score. Figure 3 provides visualizations of model predictions on slices with nodules, demonstrating its ability to precisely differentiate nodule from non-nodule regions.

The proposed model was trained and evaluated in a Google Colab environment using an L4 GPU, ensuring enough computational power for high-resolution 3D CT scans. The training was conducted over 15 epochs using the AdamW optimizer with a learning rate of 1e-4 for the main parameters and 1e-5 for the backbone parameters, combined with a weight decay of 1e-4 to reduce overfitting. The learning rate was adjusted dynamically using a Step Learning Rate Scheduler with a step size of 10 and a gamma of 0.1, which reduced the learning rate by a factor of 10 every 10 epochs to help stabilize training. The model utilized a batch size of 6 with mixed precision (16-bit floating-point), which improved training speed and efficiency. Gradient clipping was applied with a value of 0.1 to prevent exploding gradients, and the model's gradient updates were accumulated over 6 batches to stabilize learning.

\begin{table}[h]
\centering
  \caption{Performance Metrics of Deformable-DETR for Sparse Lung Tumor Detection}
  \label{tab:performance_metrics}
  \centering
  \begin{tabular}{ll}
    \toprule
    Metric                                         & Value      \\
    \midrule
    \textbf{F1 Score}                                       & \textbf{94.2\%}      \\
    \textbf{Average Precision @ IoU 0.5 (All Areas)}                 & \textbf{93.3\%}      \\
    Average Precision @ IoU 0.5 (Small Areas)               & 78.4\%      \\
    Average Precision @ IoU 0.5 (Medium Areas)              & 96.7\%      \\
    Average Precision @ IoU 0.5 (Large Areas)               & 97.8\%      \\
    \textbf{Average Recall @ IoU 0.5 (All Areas)}                 & \textbf{95.2\%}      \\
    Average Recall @ IoU 0.5 (Small Areas)               & 83.3\%      \\
    Average Recall @ IoU 0.5 (Medium Areas)              & 97.0\%      \\
    Average Recall @ IoU 0.5 (Large Areas)               & 99.2\%      \\
    \bottomrule
  \end{tabular}
\end{table}
Table \ref{tab:performance_metrics} summarizes the performance metrics for Lung-DETR on the LUNA16 test dataset. Nodules are categorized by size: small (up to 7 mm), medium (7 mm to 15 mm), and large (greater than 15 mm). Precision measures the proportion of correctly identified nodules among all predictions, while recall indicates the proportion of actual nodules detected. Average Precision (AP) at an Intersection over Union (IoU) threshold of 0.5 reflects the area under the precision-recall curve, specifically for detections with at least 50\% overlap with the ground truth, highlighting the model’s balance between precision and recall. Average Recall (AR) measures the average proportion of true positives detected across different nodule sizes. The F1 score combines precision and recall, providing a balanced evaluation of the model’s accuracy in handling false positives and negatives.

The results show that Lung-DETR achieves strong precision and recall across most tumor size bands, demonstrating its effectiveness in distinguishing between tumor and non-tumor regions despite a significant class imbalance, with only 12.7\% of the data representing the positive class. For medium and large tumors, the model maintains high precision (96.7\% and 100\%, respectively) and high recall (100\% for both), minimizing false positives, which is crucial in medical imaging to avoid unnecessary tests, procedures, and patient anxiety. Its high recall also indicates a high detection rate for actual tumors, which is vital for early diagnosis and treatment, particularly given the sparse occurrence of positive cases in the dataset.

The model shows relatively lower precision and recall for small nodules (up to 7 mm in diameter), reflecting the inherent challenges of detecting small nodules due to their lower contrast in CT scans. This also poses difficulties in real-world clinical practice. Notably,  the prevalence of malignancy in nodules smaller than 6 mm is very low, ranging between 0 and 1\%, and guidelines from the European Respiratory Society now suggest a threshold of 6 mm for follow-up consideration due to the low malignancy risk associated with these small nodules \cite{Larici170025}.

Figure \ref{fig:predvis} shows six CT slices with positive nodule regions, where green boxes denote ground truth annotations and red boxes indicate Lung-DETR’s predictions. The images reveal complex vascular structures and bronchioles that can easily mimic or obscure small nodules. Despite these complexities Lung-DETR’s predictions closely match the ground truth across all slices, even when nodules are located near dense vascular networks or airways with minimal visual contrast. The model’s consistent accuracy in detecting lung nodules and ability to detect the absence of nodules in intervening slices indicates its potential effectiveness in real world scenarios. 

This work proposes Lung-DETR, a Deformable Detection Transformer-based approach for detecting sparse lung tumors in CT scans, formulated as an anomaly detection problem to effectively manage the lung nodule sparsity present in real-world datasets. Leveraging custom preprocessing techniques, such as Maximum Intensity Projection (MIP) for enhanced 3D contextual representation, and incorporating focal loss to prioritize challenging detections, Lung-DETR achieved cutting-edge performance on the LUNA16 dataset with an F1 score of 94.2\%. The model demonstrated near-perfect precision and recall across medium and large tumor size bands, indicating its robustness in accurately distinguishing tumor regions from non-tumor regions, even in anatomically complex settings. The model’s ability to balance sensitivity and specificity shows promise for clinical applications in early lung cancer detection. Future research will aim to validate the model’s utility across a wider range of clinical datasets from different CT machines and hospitals to enhance generalizability and improve detection capabilities for small tumors, which remain a critical challenge in early diagnosis.

\bibliographystyle{plainnat}
\bibliography{references}


\newpage
\end{document}